\documentclass[11pt]{article}
\usepackage[T1]{fontenc}
\usepackage{lmodern}
\usepackage{microtype}
\usepackage{natbib}
\usepackage{enumitem}
\usepackage[a4paper, margin=1in]{geometry}
\usepackage{amsmath}
\usepackage{amssymb}
\usepackage{xcolor}
\usepackage{graphicx}
\usepackage{authblk}
\usepackage[hidelinks]{hyperref}
\usepackage[textsize=tiny]{todonotes}

\definecolor{red}{RGB}{192,0,0}
\renewcommand{\epsilon}{\varepsilon}

\title{\resizebox{0.98\textwidth}{!}{Effective Parameters, Real Behavior: Renormalization for Robotics}\\[0.6em]
\normalfont\large From Infinite Electron Mass to Sim-to-Real Gap}
\author[1,*]{Youran Sun}
\author[2]{Jiaxuan Guo}
\author[3]{Xingyu Ren}
\author[1]{Chugang Yi}
\author[1]{Haizhao Yang}
\affil[1]{University of Maryland, College Park}
\affil[2]{Stanford University}
\affil[3]{The Chinese University of Hong Kong}
\affil[*]{Corresponding author: \href{mailto:syouran0508@gmail.com}{\texttt{syouran0508@gmail.com}}}
\date{}

\begin{document}

\maketitle

\begin{abstract}

Bridging the sim-to-real gap is a central problem in robotics, and the prevailing approach is to build increasingly accurate simulators.
Here, we propose another approach based on renormalization: using effective, resolution-dependent parameters to absorb details omitted by the simulator and reproduce real behavior.
These parameters may differ from measured physical values because they compensate for what the simulator leaves out.
We demonstrate this mechanism analytically for proportional--derivative (PD) control at finite simulation frequency, where proportional feedback changes the effective derivative gain and derivative feedback changes the effective inertia.
We then interpret dynamic rope manipulation and underwater swimming through the same perspective.
Finally, we present a practical procedure for choosing observables, identifying omitted physics, and determining effective parameters.
Renormalization offers robotics a complementary path across the sim-to-real gap: effective parameters, real behavior.
\end{abstract}

\tableofcontents

\section{Introduction}

In quantum electrodynamics (QED), the bare electron mass can diverge.\footnote{
More precisely, the regulator-dependent bare mass, or its divergent contribution, can diverge.
The electron mass measured in experiments remains finite.
}
Yet QED is one of the most precisely tested theories in science and the work that established modern QED was recognized by the 1965 Nobel Prize in Physics.
How can a theory containing an infinite parameter make finite and extraordinarily accurate predictions?
The answer is renormalization \citep{peskin1995introduction}.

A physical theory cannot describe every detail of the world.
Renormalization accounts for the details left out of a model by absorbing their combined effects into a small number of parameters in the equations that remain.
These parameters can differ from quantities measured directly in the laboratory.
They are chosen so that observable behavior remains correct.
Renormalization permits wrong-looking parameters to produce the right measurable behavior.

The same idea later became useful far beyond high-energy physics (HEP).
In computational fluid dynamics (CFD), turbulent flow contains swirling structures, known as eddies, at many different sizes.
A practical simulation cannot resolve the smallest eddies.
However, in large-eddy simulation (LES), their effect on the resolved flow is represented through an effective viscosity and a small number of model coefficients.
The coarse simulation therefore uses a grid-dependent effective viscosity in addition to the measured molecular viscosity.
LES has become a standard tool in engineering fluid mechanics \citep{sagaut2006large}.

Robot simulators face the same structural problem.
They use finite time steps and simplified models of actuators, deformable objects, contact, and surrounding fluids.
The usual instinct is to measure inertia, damping, stiffness, friction, and other physical properties as accurately as possible, then copy those values into the simulator.
Renormalization suggests another possibility: the simulator may reproduce the real robot motion more accurately when some of these parameters $\theta$ are deliberately moved away from their measured values
\begin{equation}
\theta_{\mathrm{sim}}^\star(a)
\neq
\theta_{\mathrm{measured}},
\qquad
\mathcal O_{\mathrm{sim}}
\!\left(a,\theta_{\mathrm{sim}}^\star(a)\right)
\approx
\mathcal O_{\mathrm{real}}.
\end{equation}
Here, \(a\) denotes the chosen simulation resolution, such as the physics time step or the length of one segment in a discretized rope model; and $\mathcal O$ denotes the observable behavior to be matched between simulation and reality.
The parameters \(\theta_{\mathrm{sim}}^\star(a)\) absorb the effects of details that the simulator does not represent explicitly.
Could part of the sim-to-real gap be filled by using the renormalized parameters $\theta_{\mathrm{sim}}$ (which are physically incorrect but behaviorally correct) instead of the physically measured parameters $\theta_{\mathrm{measured}}$ under the philosophy of renormalization?

We illustrate this argument through three examples.
First, we show analytically that finite-frequency simulation mixes proportional feedback \(K_P\) into the effective damping \(K_D\), and damping into the effective inertia \(J\).
Second, we examine dynamic rope manipulation, where a coarse rope model succeeds without reproducing the complete real trajectory.
Third, we consider an underwater robot whose complex interaction with the surrounding fluid is represented by a small number of fitted parameters.
Together, these examples suggest that robotics should devote more attention to renormalization, because making simulations ever more accurate is not the only way to bridge the sim-to-real gap.

\section{PD Control at Finite Simulation Frequency}
\label{sec:pd_control}

Many robot-learning systems use a policy to output desired joint positions \(q_d(t)\) and desired joint velocities \(\dot q_d(t)\).
A low-level proportional--derivative (PD) controller then converts these commands into joint torques \(\tau(t)\).
During reinforcement-learning (RL) training, a physics simulator repeatedly evaluates the controlled dynamics at a finite simulation frequency \(f_{\mathrm{sim}}\).
Reducing \(f_{\mathrm{sim}}\) directly reduces the number of physics steps required for each rollout and can therefore accelerate RL training.
However, reducing \(f_{\mathrm{sim}}\) also changes the dynamics represented by the simulator.

The ideal continuous-time dynamics are
\begin{equation}
J\ddot q(t)
=
\tau(t)
=
K_P\bigl[q_d(t)-q(t)\bigr]
+
K_D\bigl[\dot q_d(t)-\dot q(t)\bigr].
\label{eq:ideal_pd}
\end{equation}
Here, \(J\) is the joint inertia, while \(K_P\) and \(K_D\) are the proportional and derivative gains.
At a finite simulation frequency, the joint state used by the controller is slightly behind the continuously evolving state.
We represent this difference by a small effective delay \(\delta t\), where
\begin{equation}
\delta t\propto 1/f_{\mathrm{sim}}.
\label{eq:delay_frequency}
\end{equation}
The simulated dynamics then become
\begin{equation}
J\ddot q(t)
=
K_P\bigl[q_d(t)-q(t-\delta t)\bigr]
+
K_D\bigl[\dot q_d(t)-\dot q(t-\delta t)\bigr].
\label{eq:delayed_pd}
\end{equation}
The delayed position and velocity can be expanded as
\begin{equation}
q(t-\delta t)
=
q(t)-\delta t\,\dot q(t)
+
O(\delta t^2),
\label{eq:position_expansion}
\end{equation}
and
\begin{equation}
\dot q(t-\delta t)
=
\dot q(t)-\delta t\,\ddot q(t)
+
O(\delta t^2).
\label{eq:velocity_expansion}
\end{equation}
Substituting Eqs.~\eqref{eq:position_expansion} and \eqref{eq:velocity_expansion} into Eq.~\eqref{eq:delayed_pd} gives
\begin{equation}
J\ddot q
={}
K_P(q_d-q)
+
K_D(\dot q_d-\dot q)
\nonumber
+
\delta tK_P\dot q
+
\delta tK_D\ddot q
+
O(\delta t^2).
\label{eq:substituted_pd}
\end{equation}
Moving the acceleration correction to the left-hand side yields
\begin{equation}
\left(J-\delta tK_D\right)\ddot q
={}
K_P(q_d-q)
+
K_D(\dot q_d-\dot q)
\nonumber
+
\delta tK_P\dot q
+
O(\delta t^2).
\label{eq:rearranged_pd}
\end{equation}
Using
\begin{equation}
\dot q
=
\dot q_d-(\dot q_d-\dot q),
\end{equation}
we obtain
\begin{equation}
\left(J-\delta tK_D\right)\ddot q
={}
K_P(q_d-q)
+
\left(K_D-\delta tK_P\right)
(\dot q_d-\dot q)
+
\delta tK_P\dot q_d
+
O(\delta t^2).
\label{eq:effective_pd}
\end{equation}

Equation~\eqref{eq:effective_pd} shows three effects of finite simulation frequency.
First, the proportional gain changes the effective derivative gain
\begin{equation}
\boxed{
K_D^{\mathrm{eff}}
=
K_D-\delta tK_P.
}
\label{eq:effective_damping}
\end{equation}
Second, the derivative gain changes the effective inertia
\begin{equation}
\boxed{
J^{\mathrm{eff}}
=
J-\delta tK_D.
}
\label{eq:effective_inertia}
\end{equation}
Third, a moving desired position produces an additional velocity-feedforward term
\begin{equation}
\boxed{
\delta tK_P\dot q_d.
}
\label{eq:extra_feedforward}
\end{equation}
Finite simulation frequency moves \(K_P\) into the effective derivative gain, moves \(K_D\) into the effective inertia, and adds a small velocity-feedforward term.
Appendix~\ref{app:effective_action} gives a complementary multi-joint effective-action derivation of these three corrections.

We can now ask what parameters should be entered into the simulator if its dynamics are required to match those of the real robot.
Let \(K_P^{\mathrm{real}}\) and \(K_D^{\mathrm{real}}\) denote the gains used on the real robot, let \(J^{\mathrm{measured}}\) denote its measured inertia, and let \(K_P^{\mathrm{sim}}\), \(K_D^{\mathrm{sim}}\), and \(J^{\mathrm{sim}}\) denote the parameters entered into the simulator.
Matching the proportional, derivative, and inertial coefficients requires
\begin{equation}
K_P^{\mathrm{sim}}
=
K_P^{\mathrm{real}},
\label{eq:kp_matching}
\end{equation}
\begin{equation}
K_D^{\mathrm{sim}}
-
\delta tK_P^{\mathrm{sim}}
=
K_D^{\mathrm{real}},
\label{eq:kd_matching}
\end{equation}
and
\begin{equation}
J^{\mathrm{sim}}
-
\delta tK_D^{\mathrm{sim}}
=
J^{\mathrm{measured}}.
\label{eq:j_matching}
\end{equation}
Solving these equations gives
\begin{equation}
K_P^{\mathrm{sim}} = K_P^{\mathrm{real}},
\label{eq:kp_sim}
\end{equation}
\begin{equation}
\boxed{K_D^{\mathrm{sim}} = K_D^{\mathrm{real}} + \delta tK_P^{\mathrm{real}} + O(\delta t^2),}
\label{eq:kd_sim}
\end{equation}
and
\begin{equation}
\boxed{J^{\mathrm{sim}} = J^{\mathrm{measured}} + \delta tK_D^{\mathrm{real}} + O(\delta t^2).}
\label{eq:j_sim}
\end{equation}
In particular,
\begin{equation}
\boxed{J^{\mathrm{sim}} \neq J^{\mathrm{measured}}.}
\label{eq:wrong_inertia}
\end{equation}
The measured inertia describes the physical robot.
The simulator inertia also compensates for dynamics lost at the finite simulation frequency.
A deliberately incorrect inertia can therefore produce a more faithful simulation.

The additional term \(\delta tK_P\dot q_d\) requires a small command-side correction when \(\dot q_d\neq0\).
It can be absorbed by slightly rescaling \(\dot q_d\), or by using separate derivative gains for velocity feedforward and velocity feedback.
The corrections to \(K_D\) and \(J\) derived above remain unchanged.

The dimensions of the correction terms provide a simple consistency check
\begin{equation}
[\delta tK_P]=[K_D],
\qquad
[\delta tK_D]=[J].
\label{eq:dimensional_check}
\end{equation}
Dimensional analysis fixes the form of the corrections.
The precise magnitude of \(\delta t\) depends on the simulator implementation.

The finite-frequency simulator does not resolve the continuous evolution between physics updates.
The resulting difference is absorbed into the derivative gain, the inertia, and a small command correction.
The simulator parameters therefore depend on the simulation frequency and need not equal their directly measured values.
This is the simplest example of renormalization in robotics considered in this paper.
Increasing \(f_{\mathrm{sim}}\) reduces these corrections.
Adjusting the effective parameters offers another way to reduce the gap when a higher simulation frequency is computationally expensive.

Actuator Reality Shaping \citep{yamamori2026actuator} applies a closely related idea on the hardware side.
A two-degree-of-freedom feedforward--feedback controller shapes each physical actuator to follow the ideal second-order dynamics assumed in simulation.
Our derivation changes simulator parameters to match the physical actuator, while Actuator Reality Shaping changes the physical actuator response to match the simulator.

A few gains and inertia parameters may become insufficient when compliance, friction, delay, and history dependence are all important.
\citep{hwangbo2019learning} represented this more complicated case with a learned actuator model constructed from real-robot data.

\section{Dynamic Rope Manipulation with a Simplified Model}

\begin{figure}[htbp]
\centering
\includegraphics[width=0.3\textwidth]{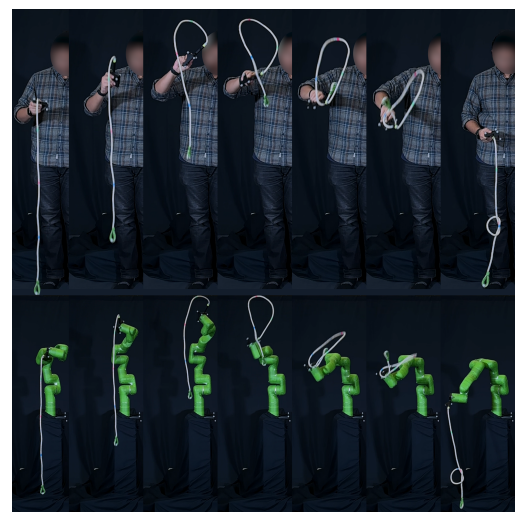}
\caption{Stages of a flying knot performed by a human and a robot. Adapted from \citet{suresh2026learning}.}
\label{fig:flying-knot-experiment}
\end{figure}

The PD example shows that a simulator can become more faithful by changing a small number of parameters.
A deformable rope presents a different challenge because its continuous shape contains far more degrees of freedom than a practical simulator can represent.
A recent study \citep{suresh2026learning} considers a robot arm that ties a flying knot by rapidly swinging a rope as shown in Figure~\ref{fig:flying-knot-experiment}.
Simulating the complete deformation and self-contact of a real rope would be expensive, so the authors replace it with a short chain of point masses connected by simple constraints.

The simplified chain does not reproduce the real rope motion accurately.
Nevertheless, it can still help the robot improve its motion.
Why can such a crude model be useful?
Tying the knot does not require the simulated and real ropes to follow the same trajectory at every instant.
It is sufficient to reproduce the behavior at the moments that determine whether the knot succeeds.

The authors therefore define the first rope--rope collision as the critical point and concentrate the correction on the rope state at that moment.\footnote{In renormalization language, matching the rope shape and velocity at the first self-collision provides the renormalization condition.}
The full trajectory can be wrong when the task-critical state is right.
Using this coarse model, the robot succeeds on all seven physically different ropes.

This example contains two reductions
\begin{equation}
\begin{aligned}
\text{continuous rope}
&\;\longrightarrow\;
\text{short point-mass chain},
\\
\text{complete trajectory}
&\;\longrightarrow\;
\text{rope state at the critical point}.
\end{aligned}
\end{equation}

Let \(x_{\infty}\) denote the state of the continuous rope, and let \(x_N\) denote the finite-dimensional state of a chain with \(N\) segments.
Let \(\theta_{\mathrm{real}}\) and \(\theta_N^\star\) denote the corresponding real and effective rope parameters.
The required agreement can then be written as
\begin{equation}
\boxed{
\mathcal O_c
\bigl(x_{\infty};\theta_{\mathrm{real}}\bigr)
\approx
\mathcal O_c
\bigl(x_N;\theta_N^\star\bigr).
}
\end{equation}
Here, \(\mathcal O_c\) denotes the rope shape and velocity at the critical point.
The continuous rope and the finite chain have different numbers of degrees of freedom and different parameters, but they preserve the same critical-point observable.

In practice, the robot uses real-world trials to adjust its command until the desired critical-point state is reached.
It does not identify every physical property or reconstruct every missing motion of the rope.
The combined effect of these missing details is absorbed into the adjusted robot command.

This example extends the renormalization argument beyond the PD controller.
A coarse simulator can succeed without reproducing the full state of reality, provided that it preserves the state that determines the task outcome.

The result also raises a quantitative question.
How many independent changes to the robot command can meaningfully change the rope state at the critical point?
This number may help determine the minimum amount of real-world experimentation required for the task.
A related study \citep{lim2022planar} uses physical cable trajectories to tune a dynamics simulator before learning a planar casting task.

\section{Underwater Swimming with a Five-Parameter Fluid Model}

\begin{figure}[htbp]
\centering
\includegraphics[width=0.4\textwidth]{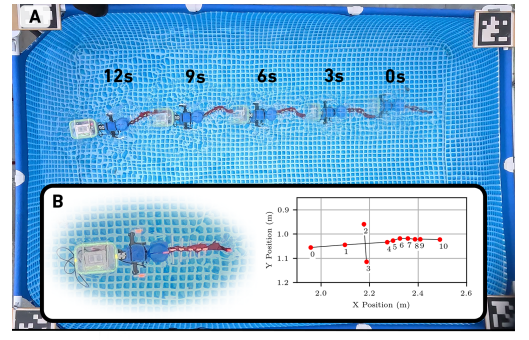}
\caption{The robotic fish swimming in a pool. Adapted from \citet{michelis2026simple}.}
\label{fig:underwater-swimmer-experiment}
\end{figure}

The rope example shows that a coarse model can be useful when it preserves a task-critical state.
An underwater robot presents an even larger reduction because its motion depends on the continuous fluid surrounding its body.
A recent study \citep{michelis2026simple} models a tendon-driven robotic fish using a simplified, stateless hydrodynamic model.
Figure~\ref{fig:underwater-swimmer-experiment} shows the swimming experiment.
Instead of simulating the velocity and pressure throughout the water, the model computes fluid forces directly from the current motion of the robot.

The authors fit five hydrodynamic coefficients using two real swimming trajectories.
The same coefficients then predict the robot's forward-swimming behavior at additional actuation frequencies that were not used during fitting.
This generalization shows that the coefficients capture a repeatable part of the fluid dynamics instead of matching only the two calibration trajectories.

Let \(x_{\mathrm{robot}}\) denote the state of the robotic fish, \(x_{\mathrm{fluid}}\) the state of the surrounding fluid, and \(\theta_5^\star\) the five fitted hydrodynamic coefficients.
The required agreement can be written as
\begin{equation}
\boxed{
\mathcal O_{\mathrm{swim}}
\bigl(x_{\mathrm{robot}},x_{\mathrm{fluid}}\bigr)
\approx
\mathcal O_{\mathrm{swim}}
\bigl(x_{\mathrm{robot}};\theta_5^\star\bigr).
}
\end{equation}
Here, \(\mathcal O_{\mathrm{swim}}\) denotes the swimming behavior to be preserved, such as the trajectory or forward speed.
The left-hand side contains the state of a continuous fluid field.
The right-hand side contains no simulated fluid state and replaces its influence on the robot with five fitted coefficients
\begin{equation}
\text{continuous fluid field}
\;\longrightarrow\;
\text{five hydrodynamic coefficients}.
\end{equation}

Viewed in this way, the five hydrodynamic coefficients are the renormalized parameters of the simplified fluid model.
Their fitted values carry the combined influence of fluid motion that is no longer simulated explicitly, allowing the model to preserve the robot's swimming behavior without representing the full fluid field.

\section{A Practical Procedure}

The examples above suggest a practical procedure for applying renormalization to a new robotics problem:

\begin{enumerate}[leftmargin=*,topsep=0.4em,itemsep=0.35em,parsep=0pt]
\item \textbf{Choose the observable behavior to preserve.}
This may be a joint response, a rope state at a critical event, a swimming speed, or another task-relevant measurement.

\item \textbf{Identify which physical details the simulator will omit at the chosen resolution.}

\item \textbf{Select the smallest set of simulator parameters that may absorb the effects of the omitted details.}
Initialize these parameters from physical measurements, but allow them to move away from those measurements during matching
\begin{equation}
\boxed{\theta_{\mathrm{measured}} \;\longrightarrow\; \theta_{\mathrm{sim}}^\star}
\end{equation}

\item \textbf{Determine the parameters from real-world measurements and validate them under new conditions.}
If the same parameters continue to preserve the chosen behavior, the omitted physics has been absorbed at the required accuracy.
Otherwise, the parameter set or the simulator must be expanded.
\end{enumerate}

This procedure translates the philosophy of renormalization into a practical strategy for sim-to-real research.

\section{Outlook}

This paper calls for greater attention to renormalization in robotics.
When a simulator omits physical details, researchers should consider whether their effects can be absorbed into a small number of effective parameters.
These parameters may depend on the simulation resolution and may differ from values measured directly on the physical system.

Several successful robotics studies already follow this logic without describing it as renormalization.
Making the connection explicit may help turn such individual insights into a systematic line of research.
For a given sim-to-real problem, researchers can ask which physical details have been omitted, which parameters can absorb their effects, which observable behavior should be preserved, and how many real-world experiments are needed to determine those parameters.

We hope that robotics will devote more effort to identifying and testing such effective parameters.
The shortest path from simulation to reality may pass through the effective parameters.

\appendix

\newpage

\section{Effective-Action Derivation for a Multi-Joint Robot}
\label{app:effective_action}

This appendix derives the multi-joint counterparts of the finite-frequency PD corrections in Section~\ref{sec:pd_control} by introducing and integrating out a history field for the controller delay.
For a robot with \(n\) joints, let \(\mathbf q(t)\in\mathbb R^n\) denote the generalized joint coordinates.
The mechanical action is
\begin{equation}
S_{\mathrm{mech}}[\mathbf q]
=
\int dt\,
L(\mathbf q,\dot{\mathbf q}),
\label{eq:mechanical_action}
\end{equation}
with
\begin{equation}
L(\mathbf q,\dot{\mathbf q})
=
\frac{1}{2}
\dot{\mathbf q}^{\mathsf T}
\mathbf M(\mathbf q)
\dot{\mathbf q}
-
V(\mathbf q).
\label{eq:robot_lagrangian}
\end{equation}
Here, \(\mathbf M(\mathbf q)\) is the joint-space inertia matrix and \(V(\mathbf q)\) is the potential energy.
Define the Euler--Lagrange operator
\begin{equation}
\boldsymbol{\mathcal E}[\mathbf q]
=
\frac{d}{dt}
\frac{\partial L}{\partial\dot{\mathbf q}}
-
\frac{\partial L}{\partial\mathbf q}.
\label{eq:euler_lagrange_operator}
\end{equation}
In standard robot-dynamics notation,
\begin{equation}
\boldsymbol{\mathcal E}[\mathbf q]
=
\mathbf M(\mathbf q)\ddot{\mathbf q}
+
\mathbf C(\mathbf q,\dot{\mathbf q})\dot{\mathbf q}
+
\mathbf g(\mathbf q).
\label{eq:robot_dynamics_operator}
\end{equation}

Let \(\mathbf q_d(t),\dot{\mathbf q}_d(t)\in\mathbb R^n\) denote the desired joint positions and velocities, and let \(\mathbf K_P,\mathbf K_D\in\mathbb R^{n\times n}\) be constant gain matrices.
The delayed PD torque is
\begin{equation}
\boldsymbol\tau_{\mathrm{PD}}(t)
=
\mathbf K_P
\bigl[
\mathbf q_d(t)-\mathbf q(t-\delta t)
\bigr]
+
\mathbf K_D
\bigl[
\dot{\mathbf q}_d(t)-\dot{\mathbf q}(t-\delta t)
\bigr].
\label{eq:multi_joint_pd}
\end{equation}
The controlled dynamics satisfy
\begin{equation}
\boldsymbol{\mathcal E}[\mathbf q]
=
\boldsymbol\tau_{\mathrm{PD}}.
\label{eq:controlled_eom}
\end{equation}
Because the delayed PD force is dissipative and history dependent, we use response fields to place the complete deterministic equations of motion inside an extended action \citep{galley2014principle}.

Introduce a history field
\begin{equation}
\mathbf y(s,t)\in\mathbb R^n,
\qquad
0\leq s\leq\delta t,
\end{equation}
satisfying
\begin{equation}
(\partial_t+\partial_s)\mathbf y(s,t)=0,
\qquad
\mathbf y(0,t)=\mathbf q(t).
\label{eq:history_constraints}
\end{equation}
Its solution is
\begin{equation}
\mathbf y(s,t)=\mathbf q(t-s),
\label{eq:history_solution}
\end{equation}
and therefore
\begin{equation}
\mathbf y(\delta t,t)=\mathbf q(t-\delta t).
\label{eq:delayed_history}
\end{equation}

Introduce response fields \(\widehat{\mathbf q}(t)\) and \(\widehat{\mathbf y}(s,t)\), together with a Lagrange multiplier \(\boldsymbol\lambda(t)\) for the boundary condition in Eq.~\eqref{eq:history_constraints}.
The extended action is
\begin{align}
S_{\mathrm{ext}}
={}& \int dt\,\widehat{\mathbf q}^{\mathsf T}
\Bigl\{\boldsymbol{\mathcal E}[\mathbf q]
-\mathbf K_P\bigl[\mathbf q_d-\mathbf y(\delta t,t)\bigr]
-\mathbf K_D\bigl[\dot{\mathbf q}_d-\partial_t\mathbf y(\delta t,t)\bigr]\Bigr\}
\nonumber\\
&+ \int dt\int_0^{\delta t}ds\,
\widehat{\mathbf y}^{\mathsf T}(\partial_t+\partial_s)\mathbf y
\nonumber\\
&+ \int dt\,\boldsymbol\lambda^{\mathsf T}
\bigl[\mathbf y(0,t)-\mathbf q(t)\bigr].
\label{eq:extended_action}
\end{align}
The first term enforces the controlled robot dynamics, the second enforces the history-field equation, and the third enforces its boundary condition.

Define the effective action by integrating out the history field and its constraint fields
\begin{equation}
e^{iS_{\mathrm{eff}}[\mathbf q,\widehat{\mathbf q}]}
=
\int
\mathcal D\mathbf y\,
\mathcal D\widehat{\mathbf y}\,
\mathcal D\boldsymbol\lambda\,
e^{iS_{\mathrm{ext}}}.
\label{eq:effective_action_definition}
\end{equation}
Integration over \(\widehat{\mathbf y}\) and \(\boldsymbol\lambda\) imposes Eq.~\eqref{eq:history_constraints}, so the remaining history field is fixed by Eq.~\eqref{eq:history_solution}.
Integrating out \(\mathbf y\) then gives
\begin{equation}
\resizebox{0.78\textwidth}{!}{$\displaystyle
S_{\mathrm{eff}} = \int dt\,\widehat{\mathbf q}^{\mathsf T}
\Bigl\{\boldsymbol{\mathcal E}[\mathbf q]
-\mathbf K_P\bigl[\mathbf q_d-\mathbf q(t-\delta t)\bigr]
-\mathbf K_D\bigl[\dot{\mathbf q}_d-\dot{\mathbf q}(t-\delta t)\bigr]\Bigr\}.
$}
\label{eq:nonlocal_effective_action}
\end{equation}

Let
\begin{equation}
D=\frac{d}{dt}.
\end{equation}
Because
\begin{equation}
\mathbf q(t-\delta t)
=
e^{-\delta tD}\mathbf q(t),
\label{eq:delay_operator}
\end{equation}
Eq.~\eqref{eq:nonlocal_effective_action} can be written as
\begin{equation}
S_{\mathrm{eff}}
=
\int dt\,
\widehat{\mathbf q}^{\mathsf T}
\left\{
\boldsymbol{\mathcal E}[\mathbf q]
+
(\mathbf K_P+\mathbf K_DD)
e^{-\delta tD}\mathbf q
-
\mathbf K_P\mathbf q_d
-
\mathbf K_D\dot{\mathbf q}_d
\right\}.
\label{eq:operator_effective_action}
\end{equation}

Expanding the delay operator,
\begin{equation}
e^{-\delta tD}
=
1-\delta tD
+\frac{\delta t^2}{2}D^2
-\frac{\delta t^3}{6}D^3
+\cdots,
\label{eq:delay_expansion}
\end{equation}
gives
\begin{align}
(\mathbf K_P+\mathbf K_DD)e^{-\delta tD}
={}&
\mathbf K_P
+
(\mathbf K_D-\delta t\mathbf K_P)D
\nonumber\\
&+
\left(
-\delta t\mathbf K_D
+
\frac{\delta t^2}{2}\mathbf K_P
\right)D^2
\nonumber\\
&+
\left(
\frac{\delta t^2}{2}\mathbf K_D
-
\frac{\delta t^3}{6}\mathbf K_P
\right)D^3
+\cdots.
\label{eq:kernel_expansion}
\end{align}
Keeping the terms through first order in \(\delta t\) yields
\begin{equation}
\resizebox{0.88\textwidth}{!}{$\displaystyle
\boxed{\bigl[\mathbf M(\mathbf q)-\delta t\mathbf K_D\bigr]\ddot{\mathbf q}
+\mathbf C(\mathbf q,\dot{\mathbf q})\dot{\mathbf q}+\mathbf g(\mathbf q)
=\mathbf K_P(\mathbf q_d-\mathbf q)
+(\mathbf K_D-\delta t\mathbf K_P)(\dot{\mathbf q}_d-\dot{\mathbf q})
+\delta t\mathbf K_P\dot{\mathbf q}_d+O(\delta t^2).}
$}
\label{eq:multi_joint_effective_eom}
\end{equation}
This is the multi-joint form of Eq.~\eqref{eq:effective_pd}.

Matching the simulated and real coefficients gives
\begin{equation}
\mathbf K_P^{\mathrm{sim}}
=
\mathbf K_P^{\mathrm{real}},
\label{eq:matrix_kp_match}
\end{equation}
\begin{equation}
\mathbf K_D^{\mathrm{sim}}
-
\delta t\mathbf K_P^{\mathrm{sim}}
=
\mathbf K_D^{\mathrm{real}},
\label{eq:matrix_kd_match}
\end{equation}
and
\begin{equation}
\mathbf M^{\mathrm{sim}}(\mathbf q)
-
\delta t\mathbf K_D^{\mathrm{sim}}
=
\mathbf M^{\mathrm{real}}(\mathbf q).
\label{eq:matrix_m_match}
\end{equation}
Therefore,
\begin{equation}
\boxed{
\mathbf K_D^{\mathrm{sim}}
=
\mathbf K_D^{\mathrm{real}}
+
\delta t\mathbf K_P^{\mathrm{real}}
+
O(\delta t^2),
}
\label{eq:matrix_kd_sim}
\end{equation}
and
\begin{equation}
\boxed{
\mathbf M^{\mathrm{sim}}(\mathbf q)
=
\mathbf M^{\mathrm{real}}(\mathbf q)
+
\delta t\mathbf K_D^{\mathrm{real}}
+
O(\delta t^2).
}
\label{eq:matrix_m_sim}
\end{equation}
Equation~\eqref{eq:matrix_m_sim} is a joint-space effective inertia correction and need not correspond to a unique set of link inertial parameters.
For a moving target, complete command matching can be obtained by rescaling \(\dot{\mathbf q}_d\), or by separating velocity feedforward and velocity feedback gains.

At second order, the effective inertia becomes
\begin{equation}
\mathbf M^{\mathrm{eff}}
=
\mathbf M
-
\delta t\mathbf K_D
+
\frac{\delta t^2}{2}\mathbf K_P,
\label{eq:second_order_inertia}
\end{equation}
and the first new higher-derivative term is proportional to
\begin{equation}
\frac{\delta t^2}{2}
\mathbf K_D
\mathbf q^{(3)}.
\label{eq:jerk_term}
\end{equation}
The main text retains the first-order terms, which can be absorbed into the existing PD gains, the joint-space inertia, and the desired-velocity command.

\bibliographystyle{apalike}
\bibliography{main}

\end{document}